\documentclass[onecolumn,times,authoryear,preprint]{elsarticle}

\usepackage{prletters}
\usepackage{framed,multirow}

\usepackage{amsmath}
\usepackage{amssymb}
\usepackage{latexsym}

\usepackage{url}
\usepackage{xcolor}
\definecolor{newcolor}{rgb}{.8,.349,.1}

\usepackage[capitalise]{cleveref}

\newcommand{\R}{\mathbb{R}}

\journal{Pattern Recognition Letters}

\begin{document}

\setcounter{page}{1}

\begin{frontmatter}

\title{Additive decomposition of one-dimensional signals using Transformers}

\author[1]{Samuele \surname{Salti}\corref{cor1}} 
\cortext[cor1]{Corresponding author.}
\ead{samuele.salti@unibo.it}
\author[]{Andrea \surname{Pinto}}
\ead{pintoandrea097@gmail.com}
\author[2]{Alessandro \surname{Lanza}}
\ead{alessandro.lanza2@unibo.it}
\author[2]{Serena \surname{Morigi}}
\ead{serena.morigi@unibo.it}

\affiliation[1]{organization={Department of Computer Science and Engineering},
                addressline={Viale Risorgimento, 2}, 
                city={Bologna}, 
                citysep={},
                postcode={40136}, 
                state={BO},
                country={Italy}}

\affiliation[2]{organization={Department of Mathematics},
                addressline={Piazza di Porta San Donato 5}, 
                city={Bologna}, 
                postcode={40126}, 
                state={BO},
                country={Italy}}
                

\begin{abstract}
One-dimensional signal decomposition is a well-established and widely used technique across various scientific fields. It serves as a highly valuable pre-processing step for data analysis. While traditional decomposition techniques often rely on mathematical models, recent research suggests that applying the latest deep learning models to this problem presents an exciting, unexplored area with promising potential.
This work presents a novel method for the additive decomposition of one-dimensional signals. 
We leverage the Transformer architecture to decompose signals into their constituent components: piece-wise constant, smooth (low-frequency oscillatory), textured (high-frequency oscillatory), and a noise component. Our model, trained on synthetic data, achieves excellent accuracy in modeling and decomposing input signals from the same distribution, as demonstrated by the experimental results.
\end{abstract}


\end{frontmatter}


\section{Introduction}
\label{sec1}

Signal decomposition (SD) approaches aim to decompose non-stationary signals into their constituents. This serves as a key preliminary step in many signal processing workflows providing useful knowledge and insight into the data and the systems that generated it. 
SD enables deeper insights and more effective decision-making across various domains like environmental science, \cite{CHEN2025120615}, 
health monitoring and biomedical signal processing, \cite{LIN2022103104}, finance, time-series and speech analysis, \cite{EMD1998}, \cite{SINGH}, \cite{ZHOU2022108155}.  
SD techniques are used in many other fields where understanding the constituent parts of a complex signal is crucial for further analysis by facilitating tasks such as removing noise or artifacts and extracting key features; see, e.g. \cite{CAI2018213} and \cite{LI20211625}.

Quite a lot of work focuses on the decomposition of a multi-component signal into amplitude‑modulated (AM) and frequency‑modulated (FM) oscillatory components; see \citep{ER2023} for a comprehensive review of the popular SD approaches into its constituent AM–FM components. However, two are the significant challenges for all these methods: noise, which is ubiquitous in most real-life signals, masking the desirable information content in the data; and the presence of components with flat sections linked by abrupt jumps, commonly known as piecewise constant signals. 

Hence, to take these challenges into account, the SD problem considered in this paper is based on the hypothesis that an observed signal $f \in \R^M$ can
be mathematically represented as an additive mixture of four components,
\begin{equation}
f = {c} + {s} + {o} + {n}\;,
\label{eq:model}
\end{equation}
with $ {c} \in\R^M$ representing a piecewise constant (also known as \emph{cartoon}) component, ${s}\in\R^M$ a smooth low-oscillatory (trend or \emph{seasonal}) component, ${o} \in\R^M$ a highly-oscillatory component, and ${n} \in\R^M$ an additive noise component which we assume to be a generic Additive White Gaussian Noise (AWGN), i.e., it contains the realization of a Gaussian-distributed $M$-variate random vector with zero-mean and covariance matrix equal to a scaled identity. 

Although this specific SD problem has been recently tackled with variational approaches \citep{CHKM2022,BOYD, GHML2024}, they suffer from some shortcomings, such as low computational efficiency and high sensitivity to hyperparameters, which are usually tuned to each test signal. Data-driven methods, and in particular deep learning models, hold the potential to overcome these limitations, since they are usually very fast and can seamlessly generalize to all signals sampled from the data-generating distribution seen at training time, without any test-time tuning. 

Our research showcases one of the earliest applications of learning techniques to the SD problem \labelcref{eq:model} in the case of more general and realistic mixtures of signals which include noise, piecewise constant, and oscillatory parts. In particular, the neural architecture chosen to solve the problem is the Transformer \citep{Transformer}, first introduced in natural language processing due to its ability to capture long-range dependencies. We speculate that such dependencies can also be important in the decomposition of a signal, since the value of the components at each time step can be estimated more effectively by having a global view on the input signal.
We refer to the proposed Trasformer-based approach for solving the Signal Decomposition problem with the TSD acronym.

The contributions of this paper are as follows:
\begin{itemize}
    \item a neural network architecture for the SD problem based on Transformers;
    \item a large synthetic dataset of signals with their ground-truth components, that we make publicly available to foster research on this important topic\footnote{to be published upon acceptance} 
    \item an experimental comparison against a state-of-the-art variational method, that highlights pros and cons of the data-driven approach to SD problems.
\end{itemize}

\section{Related work}
In the field of signal decomposition, key techniques are Fourier Decomposition Method (FDM) \citep{SINGH}, and Empirical Mode Decomposition (EMD) introduced in \cite{EMD1998}. Both break down a signal into a small number of its basic oscillatory components, Intrinsic Mode Functions and Fourier intrinsic band functions, respectively. Later other methods gained popularity, like synchrosqueezing \citep{DAUBECHIES2011243}, which is similar to EMD but uses the Wavelet Transform for improved time-frequency analysis and reconstruction of individual oscillatory modes. The most recent advancements involve Variational Mode Decomposition (VMD) \citep{DKZ2014}, which decomposes a signal into distinct band-limited components. VMD is entirely non-recursive and more resistant to noise. Variational methods are used in various works, such as a two-step method \citep{CHKM2022} that separates signals into Jump, Oscillation, and Trend components, rather than just oscillatory or non-oscillatory parts. Another example \citep{BOYD} defines masked proximal operators of each of the component loss and obtains the decomposition by minimizing the sum of each component's loss. 
In general, variational methods obtain good results but require an expensive per-signal tuning since they are highly sensitive to change in the free parameters. 
An automatic selection of the free parameters based on whiteness and autocorrelation is introduced in the Predictor-Corrector SD approach proposed in \cite{GHML2024}. Another limitation of variational methods is their low computational efficiency.

From the perspective of a deep learning approach to signal decomposition, 
to our knowledge, there is only a recent proposal in \cite{ZHOU2024110670}
to compute non-stationary
decomposition of signals.
The authors present a convolutional neural network with a residual structure.  However, their additive decompositions focus only on the simpler problem of separating noisy and simple oscillatory components.

Transformer networks have become incredibly popular in signal analysis, and their attention mechanism is indeed the core reason for this success \citep{ijcai2023p759}. Despite their capabilities, Transformers remain largely unexplored for signal decomposition. This is a missed opportunity, as decomposing time series into trend and seasonal elements can greatly enhance the accuracy of tasks such as time series forecasting. Autoformer in place of the Transformers is proposed in \cite{wu2021autoformer} for long-term time series forecasting; it introduces an Auto-Correlation mechanism in place of self-attention. In Frequency Enhanced Decomposed Transformer (FEDformer) \cite{FED}, a frequency enhanced method is introduced to compute attention. However, in all these methods, while the machine learning model is used to predict future samples in the series based on the decomposed components, the decomposition itself is not tackled with a data-driven approach. Contrary to what we investigate in this paper, decomposition in these methods is still based on model-based approaches and hand-crafted rules.

\section{Proposed TSD method}

The neural architecture that we investigate in the context of additive SD is the Transformer \citep{Transformer}, first introduced in natural language processing because of its ability to capture long-range dependencies. 
While initially developed for natural language processing, the underlying principles of the transformer architecture, particularly attention, are highly versatile. This has allowed them to be successfully adapted to various data analysis applications beyond text,  including image processing, \cite{ViT}, audio and speech processing, etc..

The ability to compare different parts of the input signal seems conducive to signal decomposition, where the presence and profile of a component is suggested also to a human observer by comparisons among sections of the input. 
While the original proposal features both an encoder and a decoder, it has been quite common to use only one of the two in subsequent works, and we have opted to use only the encoder component of the original Transformer architecture. Its ability to transform an input sequence into a processed sequence of the same length seems a good fit to model the signal decomposition (SD) problem.
We only need to insert simple layers before and after the Transformer encoder to adapt it to the SD task, which are described below. We note here that also the decoder part of the architecture could be used here, to cast the SD problem as a generative task. We leave the exploration of such research path to future work.

The model takes as input a sequence of scalars $f_i, \; i=1,\dots, M$ and produces four sequences of the same length $c_i,s_i,o_i,n_i$, corresponding to the piecewise constant, smooth, oscillatory and noise components of the input sequence. 
The Transformer encoder takes as input a sequence of vectors $y_j \in \mathbb{R}^D, j=1,\dots,L$ called tokens and produces a sequence of processed tokens of the same length and dimensionality $z_j \in \mathbb{R}^D, j=1,\dots,L$. The encoder is a stack of $N$ identical layers. Each layer consists of two parts: multi-head self-attention and a fully connected feed-forward network. Each part is surrounded by a skip connection and followed by layer normalization \citep{ba2016layernormalization}. If we stack the input tokens as rows of a matrix $Y \in \R^{L\times D}$, the self-attention operator of $Y$ is
\begin{align}
     \mathrm{SA}(Y) & = \mathrm{softmax} \left (\frac{Y Q K^T Y^T}{\sqrt{D}} \right ) (Y V) 
\end{align}
where $Q \in \R^{D \times D_k}, K \in \R^{D \times D_k}, V \in \R^{D \times D_v}$ are learnable projection matrices that compute the so called queries, keys, and values out of the input matrix $Y$, the $\mathrm{softmax}$ operation is applied row-wise, and $D_k$, $D_v$ are free hyperparameter, which set the length of queries/keys and values, respectively. 
Multi-head self-attention uses $h$ $\mathrm{SA}$ modules, without sharing parameters between them and setting $D_k = D_v = D/h$, then concatenates their output along rows, and applies a final learnable linear projection $O \in \R^{D \times D}$
\begin{align}
     \mathrm{MHSA}(Y) & = \left [SA_1(Y) \mathrel{|} SA_2(Y) \mathrel{|} \dots \mathrel{|} SA_h(Y) \right ] O .
\end{align}
The feed-forward network is a 2-layer fully connected network with ReLU activation function
\begin{align}
    \mathrm{FF}(Y) & = \mathrm{ReLU}( Y W_1^T + b_1) W_2^T + b_2
\end{align}
where $W_1 \in \R^{4D \times D}, W_2 \in \R^{D \times 4D}, b_1 \in \R^{4D}, b_2 \in \R^{D}$ are learnable parameters.

In our experiments, $L$ is always less than or equal to $M$, since both the time and space complexity of the transformer blocks is quadratic in $L$. Therefore, to feed our input into the transformer encoder, we have to adjust the depth of the signal to let it become $D$, as well as to downsample the signal to create less input tokens when $L<M$.
To downsample, we divide the signal into non-overlapping chunks of size $S=M/L$.  In particular, as we will illustrate in Section \ref{sec:as}, we have evaluated the following alternatives for splitting the signal into chunks and projecting the input to the desired dimension $D$:
\begin{description}
    
    \item[no chunks] i.e., $L=M$. We project input scalars to dimension $D$ with a 1D convolution with kernel size $3$ and $D$ output channels.

    \item[sum] We project input scalars to dimension $D$ with a 1D convolution with kernel size $3$ and $D$ output channels. We then sum the $D$-dimensional embeddings of the $S$ samples in a chunk to compute the $L$ input tokens.

    \item[cat] We project input scalars to dimension $D/S$ with a 1D convolution with kernel size $3$ and $D/S$ output channels. We then concatenate the $(D/S)$-dimensional embeddings of the $S$ samples in a chunk to create the $L$ input tokens with dimension $D$.

    \item[conv] We reshape the input signal to have $L$ samples and $S$ channels and project it to dimension $D$ with a 1D convolution with kernel size $3$, $S$ input channels and $D$ output channels. 
\end{description}

\begin{figure}[th!]
    \centering
    \includegraphics[width=0.8\linewidth]{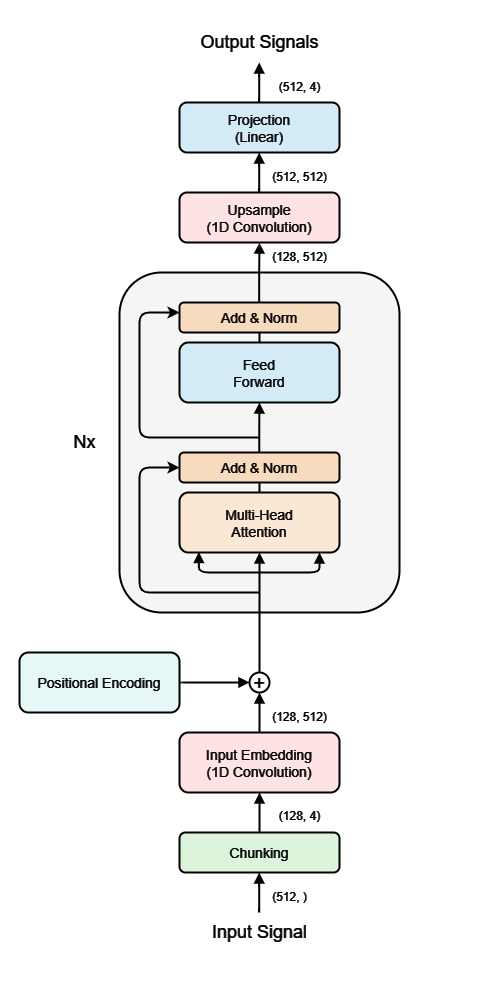}
    \caption{Neural architecture of the proposed TSD approach}
    \label{fig:architecture}
\end{figure}

To map the output tokens computed by the transformer encoder back into four signals with $M$ scalars each, we treat the sequence of tokens as a signal with $D$ samples and $L$ channels and use a 1D convolution with $M$ output channels. Then, we apply a shared linear layer on each of the $M$ samples to compute $4$ output channels, i.e., the $4$ output signals of the decomposition, from the $D$ channels computed by the network.
\cref{fig:architecture} describes our architecture with the 1D conv input variant. In particular, the figure shows the architecture with $M=512$, $D=512$,  and $L=128$, hence $S=4$.

Since Transformers are equivariant with respect to the order of the input tokens, but the SD problem is not, we add sinusoidal positional encondings \citep{Transformer} immediately after having performed downsampling and having projected the input samples to $D$ channels, to let the network be aware of the absolute position of individual tokens within the input sequence. To regularize the model, the output of this operation as well as the output of every multi-head self-attention and of every linear layer in the feed-forward modules are subject to dropout \citep{dropout} at training time.

To train the network, we minimize the sum of the Mean Square Error (MSE) for each component of a training signal, i.e. the loss for the generic signal $f \in \R^M$ that can be decomposed as $\,f=c+s+o+n\,$ is given by 
\begin{align}
    \mathcal{L} \left(c,s,o,n,\hat{c},\hat{s},\hat{o},\hat{n}\right) & \;= \!\!\sum\limits_{x \,\in\, \{c,s,o,n\}} \!\!\mathrm{MSE}\left(x, \hat{x}\right)\,,\\
    \mathrm{with}\;\;\:\mathrm{MSE}\left(x, \hat{x}\right) &\: := \frac{1}{M} \left\| x-\hat{x} \right\|_2^2,
    \label{eq:MSE}
\end{align}
where $x$ denotes the true component and $\hat{x}$ indicates the prediction of the network. 

\section{Dataset generation}
Experiments have been performed on synthetically generated datasets of signals. More precisely, each observation $f \in \R^M$ is obtained as formalized in \eqref{eq:model}, in particular
\begin{equation}
f = c + s + o + n = b_c \bar{c} + b_s \bar{s} + b_o \bar{o}+n=\bar{f} + n \, ,
\label{eq:decb}
\end{equation} 
where both the normalized (zero-mean and unitary-variance) versions $\bar{c},\bar{s},\bar{o}$ of the cartoon/smooth/oscillatory components $c,s,o$, are generated pseudo-randomly and their \emph{blending factors} $b_c,b_s,b_o \in [0,1]$ are chosen from a predefined list.

The smooth and oscillatory components $\bar{s}$ and $\bar{o}$ are generated in a similar way based on their discrete Fourier transform 
\begin{eqnarray}
\!\!\!\!\!\!\!\!\!\!\!\!\!x_i
&\!\!\!\!{=}\!\!\!\!&
\sum_{k \in \mathcal{K}}
(\alpha_k \cos(2\pi \omega_k i)+
\beta_k \sin(2\pi \omega_k i)),  
\;\:i=0,\ldots , M-1,
\label{eq:stcomp_1}\\
\!\!\!\!\!\!\!\!\!\!\!\!\!\mathcal{K}
&\!\!\!\!{=}\!\!\!\!&
\Big\{ \, k_1,\ldots,k_n \in \big[\,k_{min},k_{max} \, \big], \; n \in \big[\,n_{min}, n_{max}\,\big] \,\Big\},
\label{eq:stcomp_2}
\end{eqnarray}
with frequencies $\omega_k = k / M$ and where $n,k_i$ are generated randomly from uniform discrete distributions, whereas $\alpha_k,\beta_k$ from a uniform pdf in $[0,1]$. 
In particular, the Fourier coefficients of $\bar{s}$ and $\bar{o}$ are generated pseudo-randomly for different frequency-bands (low frequencies for $\bar{s}$, high-frequencies for $\bar{o}$). 

The piecewise constant component $\bar{c}$ is characterized in terms of the location and amplitude of its jumps (discontinuities). In particular,  the component is generated, from left to right, by selecting for each new jump its distance \mbox{$d \in [d_{min},d_{max}]$} from the previous one (or from the first sample, for the first jump) and its amplitude $a = \mathrm{sign}(a) |a|$, $\mathrm{sign}(a) \in \{-1,+1\}$, $|a| \in [a_{min},a_{max}]$. All variables $d$, $\mathrm{sign}(a)$ and $|a|$ are uniformly sampled in their domains.

Finally, the noise component $n \in \R^M$ is generated by sampling from a Gaussian pdf of $M$ variables with zero mean and diagonal covariance matrix $\sigma_n \mathrm{I}_M$ with standard deviation $\sigma_n$ that produces a prescribed value of the SNR of the observation $f$. In detail, the signal-to-noise ratio (SNR) of the noisy observation $f = \bar{f} + {n}$ is given by 
\begin{equation}
\label{eq:SNR}
\mathrm{SNR}\left(f;\bar{f}\right) = 
10\log_{10}\frac{
\left\| \bar{f}-\mathbb{E}\left[\bar{f}\right]  \right\|_2^2}
{\left\|\,n\,\right\|_2^2}\,.
\end{equation}
It follows that by corrupting $\bar{f}$ with the realization $n$ of additive zero-mean Gaussian noise of variance $\sigma_n^2$, then $\left\|n  \right\|_2^2 \simeq M \sigma_n^2$, hence choosing
\begin{equation}
\sigma_n = 10^{-\mathrm{SNR}/20}  \frac{\left\| \, \bar{f}-\mathbb{E}[\bar{f}]\, \right\|_2}{M} \, ,
\label{eq:SNR_inv}
\end{equation}
leads to an observation $f$ characterized by the prescribed SNR.

\section{Experiments}
In our experiments, the input size is $M=512$. Based on our ablation study reported in Section \ref{sec:as}, we consider two variants of our architecture: \\
i) we set the number of input tokens $L=M=512$, a variant that we refer to as \emph{TSD no chunks}; \\
ii) we set $L=128$, i.e. the chunk size $S=4$, and use the conv input layer to create the input sequence, a variant that we refer to as \emph{TSD chunks}. \\
The encoder stack is always composed of $N=4$ layers. We use the standard values for the inner dimensions $D=512$, the number of heads $h=8$, and dropout probability $p=0.1$. We train our model with ADAM \citep{KingBa15} with default hyperparameters for $15k$ epochs, with learning rate $0.0001$ and batch size $64$. We run all our experiments on a machine equipped with an Nvidia GeForce GTX 1080 with 11GB of vRAM GPU. Training takes about $48$h.

In all our experiments, we generated signals according to the following setup. For $\bar{s}$ and $\bar{o}$ components, in formulas \eqref{eq:stcomp_1}-\eqref{eq:stcomp_2} we used $[n_{min},n_{max}] = [1,3]$ for both, whereas   $[k_{min},k_{max}] = [2,7]$ for $s$ and $[k_{min},k_{max}] = [70,80]$ for $\bar{o}$.
For $\bar{c}$, we used $[d_{min},d_{max}] = [40,50]$ and $[a_{min},a_{max}] = [0.5,1]$. The blending factors in \eqref{eq:decb} for the data set of the 13 signals used to compare the variational method VSD with the proposed TSD approach have been generated to represent a broad range of component mixtures. Specifically, $[b_c,b_s,b_o]=
[1 \, 0 \, 0;\, 
0 \, 1  \, 0;\,
0 \, 0  \, 1;\,
1/3\,\, 2/3 \,\, 0;\,
2/3 \,\,1/3 \,\, 0;\,
0 \, 1/3 \,\, 2/3;\,
0 \, 2/3 \,\, 1/3; \\
1/3 \,\, 0   \,\, 2/3;\,
2/3 \,\, 0   \,\, 1/3;\,
1/3 \,\, 1/3 \,\, 1/3;\,
0.2 \,\, 0.2 \,\, 0.6;\,
0.6 \,\, 0.2 \,\, 0.2;\,\\
0.2 \,\, 0.6 \,\, 0.2].$
Finally, for what regards the noise component $n$, we used \mbox{$\mathrm{SNR} = 20$ in \eqref{eq:SNR_inv}.}

We measure the quality of a decomposed component $\hat{x} \in \R^M$ by means of the Root Mean Square Error (RMSE), given by the square root of the MSE defined in \eqref{eq:MSE}.

\subsection{Ablation study}
\label{sec:as}
To understand the impact of ours design decisions and define the best training recipe for our model, we performed an ablation study on smaller models with $N=2$ layers in the encoder, by using a reduced dataset composed of 2000 training samples, 500 validation samples and 500 test samples. While training, we save the model with the lowest average RMSE across components on the validation set, whose results on the unseen test set are then reported in Table \ref{tab:ablation}. 

The first four rows compare the different strategies to create chunks of the input signal, i.e. no chunks, sum, cat and conv, while keeping everything else constant. We can see how avoiding to downsample the signals gives the better performance on average. Interestingly, it is also the best at predicting the $c,s$ and $n$ component, while it is the worst at $o$, the highly-oscillatory one. The best on $o$, and the second best configuration on average is the conv strategy, with $S=4$. Since it also enjoys the additional advantage of requiring significantly less memory to perform a forward pass due to the quadratic complexity of Transformers with respect to the sequence length, we consider both variants in the reminder of the ablation.

Rows 5-6 study the effect of the chunk size $S$ for the conv variant. 4 turns out to be the best value.

Finally, rows 7-10 study the effect of two tentative improvements in the training recipe. In our data, in several signals one or more components are missing. Therefore the network has to sometimes output very small values for all samples of a component (ideally, all 0s). To facilitate this task, we have tried to initialize all parameters of the last linear layer responsible for computing the components to 0. This turns out to be slightly better for the no chunks variant (row 1 versus row 9), but it is not for the other one (row 4 versus 7). Hence, we adopt it in subsequent experiments only for the first one. We have also tried to add a lr scheduler \cite{1cycle} to the training recipe, which however was not effective in our experiments (rows 8 and 10).

\subsection{Comparison with variational method}
We compare the proposed TSD approach against one of the state-of-the-art variational methods presented in \cite{GHML2024}, labeled {\em VSD}. For this experiment, we use a larger dataset, comprising 18000 samples, that we split in 12000 for training, 2000 for validation and 4000 for testing. 

However, the comparison has been performed on a reduced test set of 13 samples, i.e., one random example for each blending factor used to generate the dataset. Averaged results for the four components are reported in Table \ref{tab:exp}. 
The small size of the reduced test set is due to the long, exhaustive tuning of the hyperparameters required by the variational method, which takes about 30 minutes for each signal. In fact, in general, variational methods require solving large-size optimization problems by means of iterative numerical methods for each configuration of the hyperparameters. Even if a single run of the optimization algorithm can take only a few seconds for 1D signals with hundreds of samples, the number of hyperparameter configurations to test can be very high in order to obtain optimal results.
On the other hand, data-driven methods are very fast, that is, our network in the chunks variant requires about $144 ms$ to process a batch of 128 signals on GPU and $1.6 s$ on CPU, and does not require to adjust hyperparameters to process different input signals. 

Another important practical advantage of the data-driven solution is that it is able to automatically predict the absence of a component, by returning very low values for all its samples. 
With variational methods, users have to manually specify the components to be estimated, typically after a visual examination of the input signal.

The results demonstrate the effectiveness of data-driven methods, and in particular of the proposed architecture, for the SD problem considered. The decompositions produced by our approach are more accurate then the output of VSD in all components, while being orders of magnitude faster to run and requiring no per-signal tuning. Indeed, we can seamlessly test our method on a large test set of 4000 in-distribution signals (full set, last row), obtaining similar results, which demonstrate the generalization ability of the model. The best variant in terms of raw performance is \emph{TSD no chunks}, which is however more costly to run in terms of memory and time consumption and worse in estimating the $o$ component, as seen in the ablation study. Indeed, due to the quadratic complexity of attention, the version with no chunks requires about $4$ GB of VRAM to process a batch of 128 signals, while the version with chunk size $4$ requires about $1.5$ GB to process the same mini-batch. Processing time is 400 ms without chunks and 144ms with chunks. Therefore, both variants can be useful in different application scenarios. To get the best performance regardless of resource consumption, we can also create an ensemble of the two, by using the predicted $\hat{o}$ from the chunked version and the other components from the other network.

\begin{figure*}
\centering
\begin{tabular}{c}
\includegraphics[width=15.5cm]{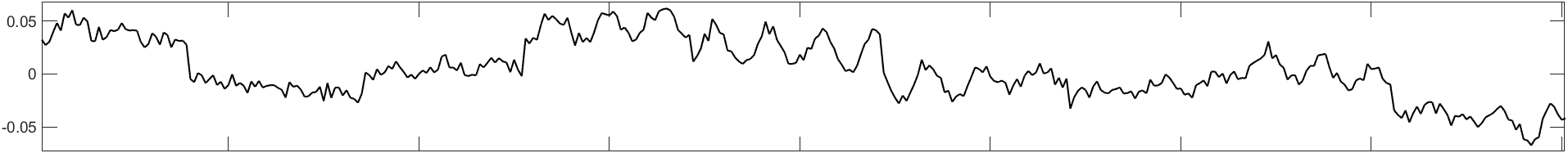}\\
\includegraphics[width=15.5cm]{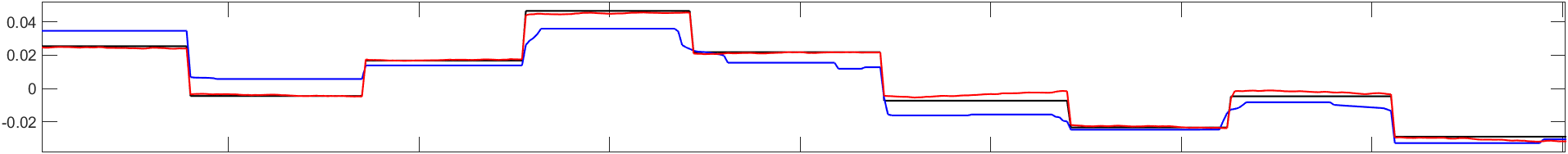}\\
\includegraphics[width=15.5cm]{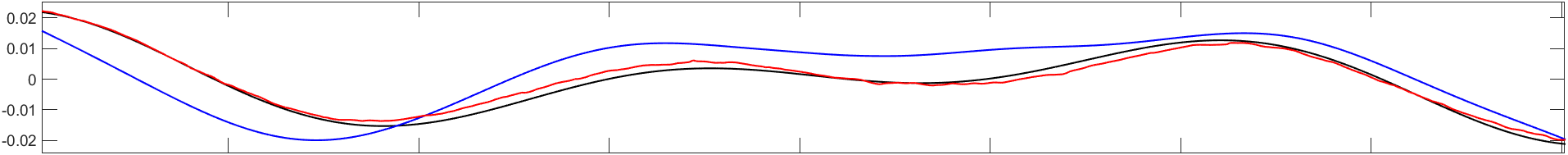}\\
\includegraphics[width=15.5cm]{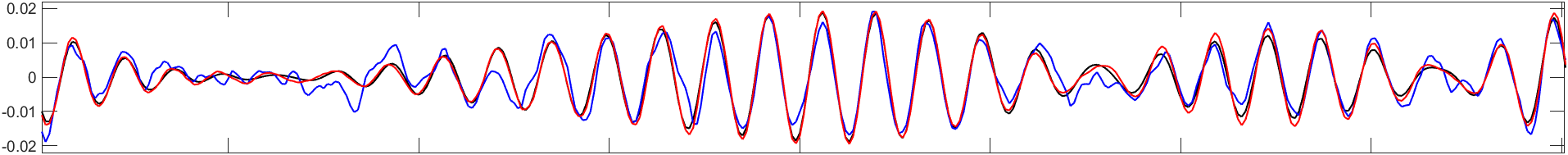}\\
\includegraphics[width=15.5cm]{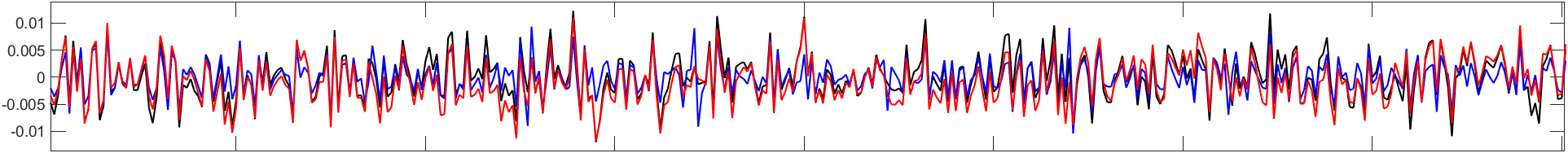}
\end{tabular}
\caption{Qualitative results on a signal from the test set: ground truth signals (black), signals estimated by VSD method (blue) and the proposed TSD data-driven approach (red). The quality metrics RMSE $\times 10^{-3}$ associated with the estimated components $\hat{c}$, $\hat{s}$, $\hat{o}$, $\hat{n}$ by VSD and the proposed TSD approach are $7.615, 7.466, 3.082, 2.395$ and $1.852, 1.558, 0.963, 2.013$, respectively.}
\label{fig:quali_signal_747}
\end{figure*}

\begin{figure*}
\centering
\begin{tabular}{c}
\includegraphics[width=15.5cm]{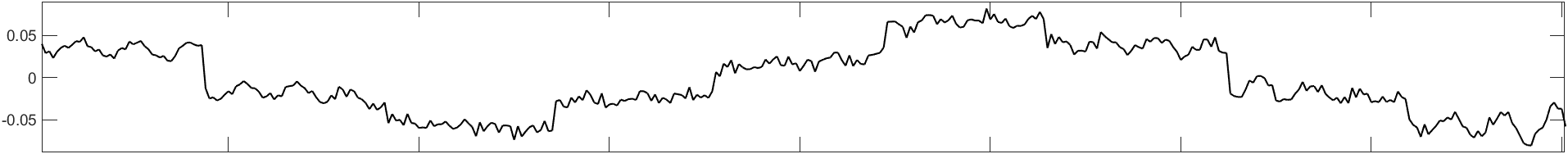}\\
\includegraphics[width=15.5cm]{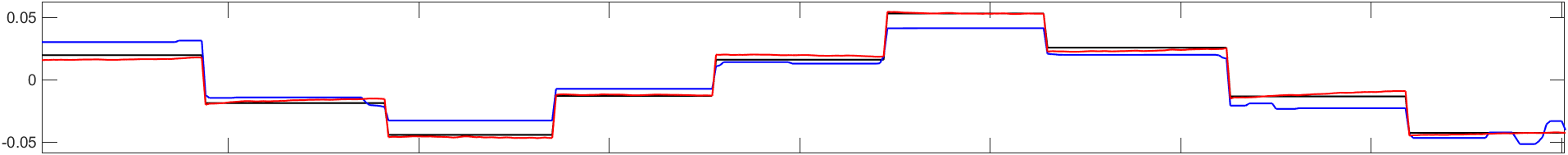}\\
\includegraphics[width=15.5cm]{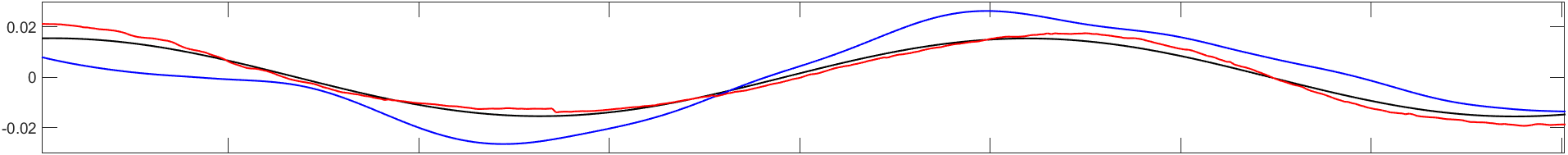}\\
\includegraphics[width=15.5cm]{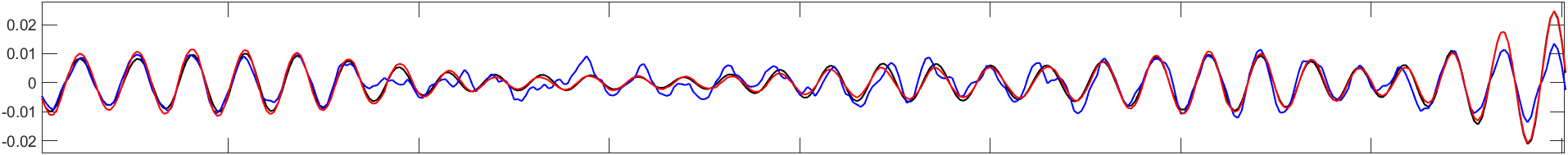}\\
\includegraphics[width=15.5cm]{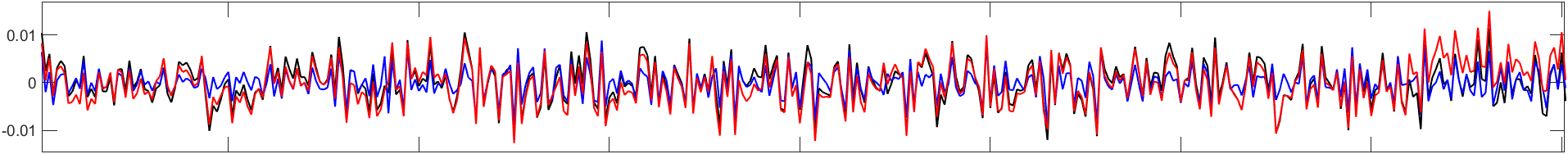}
\end{tabular}
\caption{Qualitative results on a signal from the test set: ground truth signals (black), signals estimated by VSD method  (blue) and the proposed TSD data-driven approach (red). The quality metrics RMSE $\times 10^3$ associated with the estimated components $\hat{c}$, $\hat{s}$, $\hat{o}$, $\hat{n}$ by \citep{GHML2024} and the proposed TSD approach are $8.026, 7.588, 2.899, 2.431$ and $2.317, 2.264, 0.872, 1.887$, respectively.}
\label{fig:quali_signal_695}
\end{figure*}

Finally, in Fig. \ref{fig:quali_signal_747} and Fig. \ref{fig:quali_signal_695} we show some visual results, namely the decompositions obtained on two signals from the test set. 
Both figures show in the top row the signal $f$ to be decomposed (black), whereas the ground truth signals (black) and the signals estimated by the VSD (blue) and the proposed TSD (red) approaches are depicted in the second-to-last rows. Associated RMSE values are reported in the captions.
These visual and quantitative results are coherent with those reported in Table \ref{tab:exp}, and reflect the superiority of the proposed TSD method.

\begin{table*}[]
    \centering
    \begin{tabular}{c|c|c|c|c|c|c|c|c|c}
         idx & Chunking & Chunk size & zero-init & lr scheduler & \multicolumn{5}{c}{RMSE ($\times10^{-3}$)} \\
          & & &  &  &$\hat{c}$ &  $\hat{s}$ &  $\hat{o}$ &  $\hat{n}$ & average \\
         \hline
         1& no & 1     & & & 5.996 & 5.933 & 1.840 & 2.329 & 4.024 \\
         2&sum & 4    & & & 6.654 & 5.900 & 1.211 & 4.338 & 4.526 \\ 
         3&cat & 4    & & & 7.096 & 6.164 & 1.314 & 3.667 & 4.561 \\
         4&conv & 4   & & & 6.821 & 5.572 & 1.061 & 3.546 & 4.250 \\ \hline
         5&conv & 2   & & & 6.938 & 6.133 & 1.321 & 2.838 & 4.308 \\
         6&conv & 8   & & & 6.901 & 5.994 & 1.042 & 4.364 & 4.575 \\ \hline
         7&conv & 4   & $\checkmark$ & & 6.913 & 5.722 & 1.124 & 3.577 & 4.334 \\
         8&conv & 4   & $\checkmark$ &  $\checkmark$  & 7.266 & 5.809 & 1.158 & 3.702 & 4.484 \\
         9&no & 1   & $\checkmark$ &  & 6.077 & 5.890 & 1.526 & 2.369 & 3.966 \\
         10&no & 1   & $\checkmark$ & $\checkmark$ & 6.215 & 5.796 & 1.701 & 2.531 & 4.061          
    \end{tabular}
    \caption{Ablation study. Performed on the reduced dataset with 2000 training samples, 500 validation samples and 500 test samples.}
    \label{tab:ablation}
\end{table*}

\begin{table*}[]
    \centering
    \begin{tabular}{c|c|c|c|c|c|c}
         Method & Test set &  \multicolumn{5}{c}{RMSE ($\times10^{-3}$)}\\
          & & $\hat{c}$  &  $\hat{s}$  &  $\hat{o}$  &  $\hat{n}$ & Avg\\
         \hline
         VSD  & 13 signals & 7.859 & 5.493 & 2.666 & 3.100 & 4.780\\
         TSD chunks & 13 signals & 4.248 & 3.853 & 0.879 &  3.020 & 2.999
         \\
         TSD no chunks & 13 signals & 2.983 & 2.873 & 0.997 &  1.762 & 2.153
         \\
         \hline
         TSD chunks & 4000 signals & 4.107 & 3.757 & 0.686 & 2.924 & 2.869\\
         TSD no chunks & 4000 signals & 3.230 & 3.092 & 0.843 & 1.811 & 2.244 
    \end{tabular}
    \caption{Experimental results. The first part of the table refers to a reduced test set of 13 signals. The last lines report
    the performance of our method on the full test set (4000 signals).}
    \label{tab:exp}
\end{table*}

\section{Conclusions}
In this paper, we investigated the feasibility of applying advanced deep learning techniques to additively decompose a mixture of one-dimensional non-stationary signals. We have shown that a data-driven solution can be very effective, while enjoying practical advantages like no need to tune hyperparameters and the ability to automatically detect which components are present in the input signal. We hope our  results serve as a promising foundation for future investigations into signal decomposition using transformer architectures. Potential avenues for further explorations include employing more complex and real-world datasets, extending the approach to multidimensional signals, and investigating alternative attention mechanisms to improve efficiency when processing large-scale signals.

\bibliographystyle{model5-names}
\bibliography{PAPER}
\end{document}